\documentclass{article}
\usepackage{kr}

\usepackage{times}
\usepackage{soul}
\usepackage{url}
\usepackage[hidelinks]{hyperref}
\usepackage[utf8]{inputenc}
\usepackage[small]{caption}
\usepackage{graphicx}
\usepackage{amsmath}
\usepackage{amsthm}
\usepackage{booktabs}
\usepackage{algorithm}
\usepackage{algorithmic}
\usepackage{xspace}
\usepackage{paralist}
\usepackage{microtype}
\newtheorem{example}{Example}
\newtheorem{theorem}{Theorem}
\newtheorem{definition}{Definition}
\newtheorem{lemma}{Lemma}
\newtheorem{proposition}{Proposition}

\newcommand*{\subdefref}[2]{\hyperref[{#2}]{\autoref*{#1}.\ref*{#2}}}

\usepackage{collect}
\newboolean{proofappendix}

\setboolean{proofappendix}{true}

\definecollection{proofs}
\makeatletter
\newenvironment{fullproof}[2][Proof of]
  {\@nameuse{collect}{proofs}{\begin{proof}[#1 #2]}{\end{proof}}}
  {\@nameuse{endcollect}}
\makeatother

\usepackage[inline]{enumitem}
\usepackage{stmaryrd}
\usepackage{amsfonts}
\usepackage{amssymb}
\usepackage{acronym}
\usepackage[textsize=scriptsize,disable]{todonotes}
\acrodef{BAT}{basic action theory}
\acrodef{fd-BAT}{finite-domain basic action theory}
\acrodef{MTL}{Metric Temporal Logic}
\acrodef{STS}{symbolic transition system}
\acrodef{TA}{timed automaton}
\acrodefplural{TA}{timed automata}
\acrodef{MITL}{Metric Interval Temporal Logic}
\acrodef{LTL}{Linear Time Logic}
\acrodef{ATA}{alternating timed automaton}
\acrodefplural{ATA}{alternating timed automata}
\acrodef{WSTS}{well-structured transition system}
\acrodef{qo}{quasi-ordering}
\acrodef{wqo}{well-quasi-ordering}
\acrodef{bqo}{better-quasi-ordering}

\newcommand*{\textcite}[1]{\citeauthor{#1}~(\citeyear{#1})}

\newcommand*{\titletext}{Controlling Golog Programs against MTL Constraints}

\hypersetup{
  pdftitle = {\titletext},
  pdfauthor = {Till Hofmann, Stefan Schupp},
  pdfinfo = {TemplateVersion={KR.2022.0}}
}

\title{\titletext}

\author{%
Till Hofmann$^1$\and
Stefan Schupp$^2$ \\
\affiliations
$^1$Knowledge-Based Systems Group, RWTH Aachen University, Aachen, Germany\\
$^2$Cyber-Physical Systems Group, TU Wien, Vienna, Austria\\
\emails
hofmann@kbsg.rwth-aachen.de,
stefan.schupp@tuwien.ac.at
}

\newcommand*{\golog}{\textsc{Golog}\xspace}

\newcommand*{\tacos}{\textsc{TACoS}\xspace}

\newcommand*{\es}{\texorpdfstring{\ensuremath{\mathcal{E \negthinspace S}}\xspace}{ES\xspace}}
\newcommand*{\esg}{\texorpdfstring{\ensuremath{\mathcal{E \negthinspace S \negthinspace G}}\xspace}{ESG\xspace}}
\newcommand*{\tesg}{\texorpdfstring{\ensuremath{\operatorname{\mathit{t-}}\negthinspace\mathcal{E \negthinspace S \negthinspace G}}\xspace}{t-ESG\xspace}}
\newcommand*{\mtl}{\ac{MTL}\xspace}

\DeclareMathOperator{\poss}{Poss}
\newcommand*{\eqdef}{\ensuremath{:=}}

\newcommand*{\equivspace}{\ensuremath\,\equiv\;}

\newcommand*{\until}[1]{\ensuremath{\,\mathbf{U}_{#1}}\,}

\newcommand*{\tnext}[1]{\ensuremath{\mathbf{X}_{#1}}\if#1{\,}\fi}
\newcommand*{\tprev}[1]{\ensuremath{\mathbf{V}_{#1}}\if#1{\,}\fi}
\newcommand*{\fut}[1]{\ensuremath{\mathbf{F}_{#1}}\if#1{\,}\fi}
\newcommand*{\past}[1]{\ensuremath{\mathbf{P}_{#1}}\if#1{\,}\fi}
\newcommand*{\glob}[1]{\ensuremath{\mathbf{G}_{#1}}\if#1{\,}\fi}
\newcommand*{\hist}[1]{\ensuremath{\mathbf{H}_{#1}}\if#1{\,}\fi}
\newcommand*{\mi}[1]{\ensuremath{\mathit{#1}}}
\newcommand*{\la}{\langle}
\newcommand*{\ra}{\rangle}
\newcommand*{\final}{\ensuremath{\mathcal{F}^w}}
\mathchardef\mhyphen="2D 

\newcommand*{\pre}{\ensuremath{\text{pre}}}
\newcommand*{\post}{\ensuremath{\text{post}}}
\newcommand*{\sac}[1]{\ensuremath{\mathit{start}\_{#1}}}
\newcommand*{\eac}[1]{\ensuremath{\mathit{end}\_{#1}}}
\newcommand*{\smid}{\ensuremath{\:\mid\:}}
\newcommand*{\realpos}{\mathbb{R}_{\geq 0}}
\newcommand*{\fluents}{\ensuremath{\mathcal{F}}\xspace}
\newcommand*{\worlds}{\ensuremath{\mathcal{W}}\xspace}
\DeclareMathOperator{\sub}{sub}
\newcommand*{\bat}{\ensuremath{\Sigma}\xspace}
\DeclareMathOperator{\fract}{fract}
\DeclareMathOperator{\abs}{abs}
\DeclareMathOperator{\reg}{reg}
\newcommand*{\symalphabet}{\ensuremath{\mathcal{V}}\xspace}
\newcommand*{\symlts}{\ensuremath{\mathcal{E}}\xspace}
\newcommand*{\symtrans}[2][t]{\ensuremath{\xrightarrow[#1]{#2}}\xspace}
\newcommand*{\symstates}{\ensuremath{E}\xspace}
\newcommand*{\symstate}{\ensuremath{e}\xspace}
\newcommand*{\symtraces}{\ensuremath{\mathcal{S}_\Delta}\xspace}
\newcommand*{\symtrace}{\ensuremath{s}\xspace}
\newcommand*{\syncstates}{\ensuremath{S}\xspace}
\newcommand*{\syncstate}{\ensuremath{s}\xspace}
\newcommand*{\synctrans}[1]{\ensuremath{\xrightarrow{#1}}\xspace}
\newcommand*{\trace}{\ensuremath{z}\xspace}
\newcommand*{\lts}{\ensuremath{\mathcal{S}}\xspace}
\newcommand*{\ltsstate}{\ensuremath{S}\xspace}
\newcommand*{\disq}{\ensuremath{\mathcal{T}_\sim}\xspace}
\newcommand*{\disqstate}{\ensuremath{\sigma}\xspace}
\newcommand*{\disqstates}{\ensuremath{S_{\sim}}\xspace}
\newcommand*{\trans}[1]{\xrightarrow{#1}}
\newcommand*{\regconf}{\ensuremath{h}\xspace}
\newcommand*{\regtrans}[1]{\lhook\joinrel\xrightarrow{#1}}
\newcommand*{\detregtrans}[1]{\lhook\joinrel\xrightarrow{#1}_D}
\newcommand*{\detdisq}{\ensuremath{\mathcal{DT}_\sim}\xspace}
\newcommand*{\detstates}{\ensuremath{\mi{SW}}\xspace}
\newcommand*{\detstate}{\ensuremath{c}\xspace}
\newcommand*{\hset}{\ensuremath{\mathcal{C}}\xspace}
\newcommand{\warrow}[1][w]{\ensuremath{\xrightarrow{#1}}}
\newcommand*{\traces}{\ensuremath{\mathcal{Z}}\xspace}
\newcommand*{\fluenttrace}{\ensuremath{\psi}\xspace}
\newcommand*{\fluenttraces}{\ensuremath{\Psi}\xspace}
\DeclareMathOperator{\tw}{tw}

\newcommand*{\qo}{\ac{qo}\xspace}
\newcommand*{\wqo}{\ac{wqo}\xspace}
\newcommand*{\bqo}{\ac{bqo}\xspace}
\DeclareMathOperator{\val}{val}
\DeclareMathOperator{\plays}{plays}
\DeclareMathOperator{\clockconstraints}{g}

\DeclareMathOperator{\reset}{reset}

\DeclareMathOperator{\ztime}{time}
\DeclareMathOperator{\clockvaluations}{N}
\newcommand*{\zeroclocks}{\ensuremath{\vec{0}}\xspace}
\newcommand*{\clockset}{\ensuremath{C}\xspace}
\newcommand*{\ata}[1][\phi]{\ensuremath{\mathcal{A}\ifthenelse{\equal{#1}{}}{}{}{_{#1}}}\xspace}
\newcommand*{\ataalphabet}{\ensuremath{\Sigma_{\ata[]}}\xspace}
\newcommand*{\atalocations}{\ensuremath{L}\xspace}
\newcommand*{\atastates}{\ensuremath{Q}\xspace}
\newcommand*{\ataconf}{\ensuremath{G}\xspace}
\newcommand*{\ataconfs}{\ensuremath{\mathcal{G}}\xspace}
\DeclareMathOperator{\suc}{Succ}
\newcommand*{\controller}{\ensuremath{\mi{CR}}\xspace}

\newcommand*{\grasp}{\ensuremath{\mi{grasp}}\xspace}
\newcommand*{\cam}{\ensuremath{\mi{cam}}\xspace}
\newcommand*{\camon}{\ensuremath{\mi{cam\_on}}\xspace}

\newcommand*{\grasping}{\ensuremath{\mi{grasping}}\xspace}

\newcommand*{\acam}{\cam()}
\newcommand*{\ccam}{c_{\cam}}
\newcommand*{\phibad}{\phi_{\mi{bad}}}

\begin{document}

\maketitle

\begin{abstract}
  While Golog is an expressive programming language to control the
high-level behavior of a robot, it is often tedious to use on a real
robotic system. On an actual robot, the user needs to consider low-level
details, such as enabling and disabling hardware components, e.g., a
camera to detect objects for grasping.  In other words, high-level actions
usually pose implicit temporal constraints on the low-level platform,
which are typically independent of the concrete program to be executed.
In this paper, we propose to make these constraints explicit by
modeling them as MTL formulas, which enforce the execution of certain
low-level platform operations in addition to the main program.  Based on
results from timed automata controller synthesis, we describe a method
to synthesize a controller that executes both the high-level program and
the low-level platform operations concurrently in order to satisfy the
MTL specification.  This allows the user to focus on the high-level
behavior without the need to consider low-level operations. We present
an extension to Golog by clocks together with the required theoretical
foundations as well as decidability results.

\end{abstract}

\section{Introduction}

While \golog is an expressive language to describe high-level robot behavior, it is often tedious to use on real-world robots, as it is challenging to manage all the low-level details of the system \cite{schifferSelfMaintenanceAutonomousRobots2010,hofmannConstraintbasedOnlineTransformation2018}.
A real-world platform often poses implicit requirements that must be considered while developing a robot program.
As an example, consider a simple robot that is able to pick up objects.
From a high-level perspective, it is easy to model this, e.g., 
\grasp changes the location of the object and requires that the robot is at the same location as the object.
However, on an actual robot, much more is involved, e.g., before doing \grasp, the robot's camera must have been enabled long enough so the object detection was able to find the object.
Encoding these low-level details quickly results in a bloated robot program that is difficult to maintain.
Furthermore, it requires the awareness of the developer of low-level details, as the robot's actions may otherwise have unexpected consequences, e.g., the robot breaking the object.

In this paper, we propose to make these implicit requirements explicit.
by modelling them as constraints in \ac{MTL}, a temporal logic that allows timing constraints.
Rather than having a single detailed program, we propose that the high-level program is augmented with a \emph{maintenance program}, which takes care of the low-level details and is executed concurrently. 
This allows the developer to focus on the high-level behavior while ensuring that all constraints are satisfied.

Executing such an augmented program poses a \emph{synthesis problem}: a controller needs to decide which actions to execute and when to execute them.
In this paper, we investigate this synthesis problem.
Given a high-level \golog program and an \mtl specification of undesired behavior, the task is to select those actions that
\begin{enumerate*}[label=(\roman*)]
  \item are possible successors according to the \golog program,
  \item ensure that the program eventually terminates, and
  \item satisfy the specification.
\end{enumerate*}
To solve this synthesis problem, we take inspiration from \emph{timed automata controller synthesis} \cite{bouyerControllerSynthesisMTL2006}, where a timed automaton is controlled against an \mtl specification.
We present the following contributions:
\begin{enumerate*}[label=(\roman*)]
  \item the logic \tesg, which extends \golog with clocks and clock constraints to express timing constraints,
	\item a theoretical framework for controller synthesis on \golog programs given an \ac{MTL} specification,
	\item and decidability results of the controller synthesis problem for \golog programs on finite domains.
\end{enumerate*}

We start in \autoref{sec:background} by describing the background and related work, before we introduce the logic \tesg in \autoref{sec:timed-esg}.
We summarize results on \mtl decidability in \autoref{sec:mtl}, define the controller synthesis problem in \autoref{sec:control-problem} and then describe our approach in \autoref{sec:approach}.
We conclude in \autoref{sec:conclusion}.

\section{Background and Related Work}\label{sec:background}

The situation calculus \cite{mccarthySituationsActionsCausal1963,reiterKnowledgeActionLogical2001} is a logical formalism for reasoning about dynamical domains based on first-order logic.
World states are represened explicitly as first-order terms called \emph{situations}, where \emph{fluents} describe (possibly changing) properties of the world and actions are axiomatized in \acfp{BAT}.
\golog~\cite{levesqueGOLOGLogicProgramming1997,degiacomoConGologConcurrentProgramming2000} is a programming language based on the situation calculus that allows to control the high-level behavior of robots.
\es \cite{lakemeyerSemanticCharacterizationUseful2011} is a modal variant and epistemic extension of the situation calculus, where situations are part of the semantics but do not appear as terms in the language.
\esg~\cite{classenLogicNonTerminatingGolog2008} extends \es with a transition semantics for \golog programs and temporal formulas to verify the correctness of \golog programs~\cite{classenPlanningVerificationAgent2013}.

\ac{MTL} \cite{koymansSpecifyingRealtimeProperties1990} is an extension of \ac{LTL} with metric time, which allows expressions such as $\fut{\leq c}$, meaning \emph{eventually within time $c$}.
In \ac{MTL}, formulas are interpreted over \emph{timed words}, consisting of sequences of symbol-time pairs. 
Depending on the choice of the state and time theory, the satisfiability problem for \ac{MTL} becomes undecidable \cite{alurRealTimeLogicsComplexity1993}.
However, for finite words with pointwise semantics, it has been shown to be decidable \cite{ouaknineDecidabilityMetricTemporal2005,ouaknineRecentResultsMetric2008}.
We build on top of this result and describe it in more detail in \autoref{sec:ata}.

A \acf{TA}~\cite{alurTimedAutomata1999} is an automaton equipped with a finite set of clocks, whose real-timed values increase uniformly, and where a transition may reset clocks and have a guard on the clock value.
\textcite{asarinControllerSynthesisTimed1998} describe a synthesis method where the goal is to reach a certain set of (timed) goal states of a \ac{TA}, which can be seen as an \emph{internal} winning condition.
\textcite{dsouzaTimedControlSynthesis2002} synthesize a timed controller against an \emph{external} specification, where the undesired behavior is given as a \ac{TA}, assuming \emph{fixed resources}, i.e., a fixed number of clocks and a fixed resolution in timing constraints.
They do so by defining a \emph{timed game}, where a winning strategy corresponds to a control strategy, and then reducing the timed game to a classical game.
In a similar fashion, \textcite{bouyerControllerSynthesisMTL2006} describe a timed game to determine a controller for a \ac{TA} against an \mtl specification.
Our method to find a controller for a \golog program is inspired by these techniques.

Similar to the proposed approach, \textcite{schifferSelfMaintenanceAutonomousRobots2010} extend \golog{} for
self-maintenance by allowing temporal constraints using Allen's Interval Algebra
\cite{allenMaintainingKnowledgeTemporal1983}.
Related is also the work by \textcite{finziRepresentingFlexibleTemporal2005},
who propose a hybrid approach of temporal constraint reasoning and reasoning about actions based on the situation
calculus.
Based on $\mi{LTL}_f$ synthesis~\cite{degiacomoSynthesisLTLLDL2015}, \textcite{heReactiveSynthesisFinite2017} describe a synthesis method that controls a
robot against uncontrollable environment actions under resource constraints.
In contrast to this work, they do not allow metric
temporal constraints.
\textcite{viehmannTransformingRoboticPlans2021} augment a fixed action sequence (e.g., a plan) with maintenance actions to satisfy a temporal specification with timing constraints.
In contrast to our work, they do not allow \golog programs, and they use a restricted constraint language.
\textcite{hofmannControllerSynthesisGolog2021} convert a \golog program to a \ac{TA}, which allows to use \mtl controller synthesis to solve the synthesis problem.
Rather than constructing a \ac{TA}, we synthesize a controller based on a symbolic execution of the program.


\section{Timed \esg{}}\label{sec:timed-esg}
\todo{Clocked-esg?}
We start by describing \tesg, a logic that allows the specification of \golog programs with timing constraints.
It builds on top of \esg and \es, which in turn are modal variants of the situation calculus.
In contrast to an earlier version of the logic~\cite{hofmannLogicSpecifyingMetric2018}, we do not allow arbitrary expressions referring to time, as this directly leads to undecidability, even in finite contexts.
Instead, inspired by \aclp{TA}, we introduce clocks and we only allow comparisons to clock values as well as clock resets. 

The language has three sorts: \emph{object},  \emph{action}, and \emph{clock}. A special feature inherited from \es{} is the use of countably
infinite sets of \emph{standard names} for those sorts. Standard object and clock names syntactically look like constants, but are
intended to be isomorphic with the set of all objects (clocks) of the domain. In other words, standard object (clock) names can be
thought of as constants that satisfy the unique name assumption and domain closure for objects.
We assume that object standard names include the rational numbers 
as a subsort.
Action standard names
are function symbols of any arity whose arguments are standard object names, e.g., $\mi{grasp}(o)$ for picking up an object.
Again, standard action names range
over all actions and satisfy the unique name assumption and domain closure for actions. One advantage of using standard
names is that quantifiers can be understood substitutionally when defining the semantics. For simplicity, we do not
consider function symbols other than actions.

\subsection{Syntax}
\begin{definition}[Symbols of \tesg{}]
The symbols of the language are from the following vocabulary:
\begin{enumerate}
  \item variables of sort object $x_1, x_2, \ldots$, action $a, a_1, a_2, \ldots$, and clock $c, c_1, c_2, \ldots$,
  \item standard names of sort object $\mathcal{N}_O = \{ o_1, o_2, \ldots \}$, action $\mathcal{N}_A = \{ p_1, p_2, \ldots \}$, and clock $\mathcal{N}_C = \{ q_1, q_2, \ldots \}$,
  \item fluent predicates of arity $k$: $\mathcal{F}^k: \{ F_1^k, F_2^k, \ldots \}$, e.g., $\mi{Holding(o)}$; we assume this list contains the distinguished predicates $\poss$, $\reset$, and $\clockconstraints$, and
  \item rigid predicates of arity $k$: $\mathcal{G}^k = \{ G_1^k, G_2^k, \ldots \}$.    
\end{enumerate}
\end{definition}



\begin{definition}[Terms of \tesg{}]
  The set of terms of \tesg{} is the least set such that
  \begin{enumerate*}
    \item every variable is a term of the corresponding sort,
    \item every standard name is a term of the corresponding sort.
  \end{enumerate*}
\end{definition}

\begin{definition}[Formulas]
  The \emph{formulas of \tesg{}}, consisting of \emph{situation formulas} and \emph{clock formulas}, are the least set such that
  \begin{enumerate}
    \item if $t_1,\ldots,t_k$ are object or action terms and $P$ is a $k$-ary predicate symbol,
      then $P(t_1,\ldots,t_k)$ is a situation formula,
    \item if $t_1$ and $t_2$ are object or action terms,
      then $(t_1 = t_2)$ is a situation formula,
    \item if $c$ is a clock term and $r \in \mathbb{N}$\footnote{With a finite set of rational constants used for comparison in constraints, these can be scaled to integer values.}, then $c \bowtie r$ is a clock formula, where $\operatorname{\bowtie} \in \{ <, \leq, =, \geq, > \}$,
    \item if $\alpha, \beta$ are clock formulas, then $\square \alpha$ and $\alpha \wedge \beta$ are clock formulas, and
    \item if $\alpha$ and $\beta$ are situation formulas,
      $x$ is a variable, and $\delta$ is a program expression (defined below),
      then $\alpha \wedge \beta$, $\neg \alpha$, $\forall x.\, \alpha$,
      $\square\alpha$, and $[\delta]\alpha$
      are situation formulas.
  \end{enumerate}
\end{definition}

A predicate symbol with standard names as arguments is called a \emph{primitive formula}, and we denote the set of
primitive formulas as $\mathcal{P}_F$.  We read $\square \alpha$ as ``$\alpha$ holds after executing any sequence of
actions'', $[\delta] \alpha$ as ``$\alpha$ holds after the execution of program $\delta$'',
and $c < r$ as ``the value of clock c is less than r'' (analogously for $\leq, =, \geq, >$).

  A situation formula is called {\em static}  if it contains no $[\cdot]$
      or $\square$
      operators and {\em fluent} if it is static and does not mention $\poss$.


Finally, we define the syntax of \golog{} program expresions referred to by the operator $[\delta]$: 

\begin{definition}[Program Expressions]
  \[
    \delta ::= t \smid \alpha? \smid \delta_1 ; \delta_2 \smid \delta_1|\delta_2
    \smid \delta_1 \| \delta_2 \smid \delta^*
  \]
  where $t$ is an action term and $\alpha$ is a static situation formula. A
  program expression consists of actions $t$, tests $\alpha?$, sequences
  $\delta_1;\delta_2$, nondeterministic%
  \footnote{We leave out the pick operator $\pi x.\, \delta$, as we later restrict the domain to be finite, where pick can be expressed with nondeterministic branching.}
  branching $\delta_1 | \delta_2$,
  interleaved concurrency
  $\delta_1 \| \delta_2$, and nondeterministic iteration $\delta^*$.
\end{definition}

We also use the abbreviation $\mi{nil} \eqdef \top?$ for the empty program that always succeeds.

\subsection{Semantics}


\begin{definition}[Timed Traces]
  A \emph{timed trace} is a finite timed sequence of action standard names with monotonically non-decreasing time.
Formally, a trace $\pi$ is a mapping $\pi:\mathbb{N} \rightarrow \mathcal{N}_A \times \realpos$, and for every $i,j \in \mathbb{N}$ with $\pi(i) = \left(\sigma_i,t_i\right)$, $\pi(j) = \left(\sigma_j,t_j\right)$ : If $i < j$, then $t_i \leq t_j$.
\end{definition}

We denote the set of timed traces as $\mathcal{Z}$.
For a timed trace $z = \left(a_1,t_1\right) \ldots \left(a_k,t_k\right)$, we define $\ztime(z) \eqdef
t_k$ for $k > 0$ and $\ztime(\la\ra) \eqdef 0$.

\begin{definition}[World]
  Intuitively, a world $w$ determines the truth of fluent predicates, not just initially, but after any (timed) sequence of actions.
  Formally, a world $w$ is a mapping $\mathcal{P}_F \times \mathcal{Z} \rightarrow \{ 0, 1 \}$. If $G$ is a rigid
  predicate symbol, then for all $z$ and $z'$ in $\mathcal{Z}$, $w[G(n_1,\ldots,n_k),z]=w[G(n_1,\ldots,n_k),z']$.
  The set of all worlds is denoted by $\worlds$.
\end{definition}

Similar to \es{} and \esg{}, the truth of a fluent after any sequence of actions is determined by a world $w$.
Different from $\es{}$ and $\esg{}$, we require all traces referred to by a world to contain time values for each
action.
This will allow us to specify \acs{MTL} constraints on program execution traces.

To allow the program to keep track of time, we introduce \emph{clocks}, similarly to clocks in timed automata.
At each point of the program execution, each clock needs to have a value, which is determined by the \emph{clock valuation}:
\begin{definition}[Clock Valuation]
  A clock valuation over a finite set of clocks $C$ is a mapping $\nu: C \rightarrow \realpos$.
\end{definition}
We denote the set of all clock valuations over $C$ as $\clockvaluations_C$\todo{This symbol is overloaded by clock standard names. Do we even need this?}.
The clock valuation \zeroclocks denotes the clock valuation $\nu$ with $\nu(c) = 0$ for all $c \in C$.
For every $\operatorname{\bowtie} \in \{ <, \leq, =, \geq, > \}$, we write $\nu \bowtie \nu'$ if $\nu(c) \bowtie \nu'(c)$ for all $c \in C$.


We continue by define the transitions that a program may take in a given world $w$.
The program transition semantics is based on \esg~\cite{classenLogicNonTerminatingGolog2008} and extended by time and clocks.
Here, a program configuration is a tuple $(z, \nu, \rho)$ consisting of a timed trace $z$, a clock valuation $\nu$, and the remaining program $\rho$.
A program may take a transition $(z, \nu, \rho) \warrow (z', \nu', \rho')$ if it can take a single action that results in the new configuration.
In three places, these refer to the truth of clock and situation formulas (see Definition~\ref{def:tesg-truth} below).
\begin{definition}[Program Transition Semantics]\label{def:trans}
  The transition relation \warrow{} among configurations, given
  a world $w$, is the least set satisfying
  \begin{enumerate}
    \item \label{def:trans:action}
      $\la z, \nu, a \ra \warrow \la z', \nu', \mi{nil} \ra$,
      if $z' = z \cdot (a, t)$ and if there is $d \geq 0$ such that
      \begin{enumerate}
        \item $t = \ztime(z) + d$,
        \item $w, z \models \poss(a)$,
        \item $w, z, \nu + d \models \clockconstraints(a)$, and
        \item
          \[
            \nu'(c) =
            \begin{cases}
              0 &\text{ if } w, z' \models \reset(c) \\
              \nu(c) + d &\text{ otherwise }
            \end{cases}
          \]
      \end{enumerate}
    \item $\la z, \nu,\delta_1;\delta_2 \ra \warrow
      \la z \cdot p, \nu', \gamma;\delta_2 \ra$
      \\
      if $\la z, \nu, \delta_1 \ra \warrow \la z \cdot p, \nu', \gamma \ra$,
    \item $\la z, \nu,\delta_1;\delta_2 \ra \warrow \la z \cdot p, \nu', \delta' \ra$
      \\
      if $\la z, \nu, \delta_1 \ra \in \mathcal{F}^w$ and
      $\la z, \nu, \delta_2 \ra \warrow \la z \cdot p, \nu', \delta' \ra$, 
    \item $\la z, \nu,\delta_1 | \delta_2 \ra \warrow \la z \cdot p, \nu', \delta' \ra$
      \\
      if $\la z, \nu, \delta_1 \ra \warrow \la z \cdot p, \nu', \delta' \ra$
      or $\la z, \nu, \delta_2 \ra \warrow \la z \cdot p, \nu', \delta' \ra$,
    \item $\la z, \nu,\delta^* \ra \warrow \la z \cdot p, \nu', \gamma; \delta^* \ra$ if
      $\la z, \nu,\delta \ra \warrow \la z \cdot p, \nu', \gamma \ra$,
    \item $\la z, \nu,\delta_1 \| \delta_2 \ra \warrow \la z \cdot p, \nu', \delta' \|
      \delta_2 \ra$
      \\
      if $\la z, \nu, \delta_1 \ra \warrow \la z \cdot p, \nu', \delta' \ra$, and
    \item $\la z, \nu,\delta_1 \| \delta_2 \ra \warrow \la z \cdot p, \nu', \delta_1 \|
      \delta' \ra$
      \\
      if $\la z, \nu, \delta_2\ra \warrow \la z \cdot p, \nu', \delta' \ra$.
  \end{enumerate}
  The set of final configurations \final{} is the smallest set such that
  \begin{enumerate}
    \item $\la z, \nu,\alpha? \ra \in \final$ if $w, z \models \alpha$,
    \item $\la z, \nu,\delta_1;\delta_2 \ra \in \final$
      if $\la z, \nu,\delta_1 \ra \in \final$ and $\la z, \nu,\delta_2 \ra \in \final$
    \item $\la z, \nu,\delta_1 | \delta_2 \ra \in \final$
      if $\la z, \nu,\delta_1 \ra \in \final$,
      or $\la z, \nu,\delta_2 \ra \in \final$,
    \item $\la z, \nu,\delta^* \ra \in \final$, and
    \item $\la z, \nu,\delta_1 \| \delta_2 \ra \in \final$
      if $\la z, \nu,\delta_1 \ra \in \final$
      and $\la z, \nu,\delta_2 \ra \in \final$.
  \end{enumerate}
\end{definition}

Intuitively, the program may take a single transition step (\subdefref{def:trans}{def:trans:action}) if the action is possible to execute and if there is some time increment such that all clock constraints are satisfied.
In the resulting configuration, the program trace is appended with the new action $a$ and the incremented time $t$, all clock values are either incremented by the time increment, or reset if $a$ resets the clock.
Also, a program is final if there is no remaining action and if any test $\alpha?$ is successful.


By following the transitions, we obtain \emph{program traces}:
\begin{definition}[Program Traces]\label{def:program-trace}
  Given a world $w$, a finite trace $z$, and a clock valuation $\nu$, the
  \emph{program traces} of a program expression $\delta$ are defined as follows:
  \begin{align*}
    \suc_w(z, \nu, \delta) &= \{ (z', \nu', \delta') \mid \la z, \nu, \delta \ra \warrow \la z', \nu', \delta' \ra \}
  \\
    \suc_w^*(z, \nu, \delta) &= \{ (z', \nu', \delta') \mid \la z, \nu, \delta \ra \warrow^* \la z', \nu', \delta' \ra \}
    \\
    \suc_w^f(z, \nu, \delta) &= \{ (z', \nu', \delta') \mid \la z, \nu, \delta \ra \warrow^* \la z', \nu', \delta' \ra
                 \\ & \qquad
    \text{ and } \la z', \nu', \delta' \ra \in \final \}
    \\
    \|\delta\|^{z, \nu}_w &= \{ z' \mid  (z \cdot z', \nu', \delta') \in \suc_w^f(z, \nu, \delta) \}
  \end{align*}
%
\end{definition}
Intuitively, $\suc_w(z, \nu, \delta)$ describes the direct successor configurations of a given configuration, while
$\suc^*_w(z, \nu, \delta)$ denotes all reachable configurations from the current configuration, and $\suc^f_w(z, \nu, \delta)$ denotes those sequences ending in a final configuration.
We use $\|\delta\|^{z, \nu}_w$ to denote only the timed traces resulting in a final configuration.
We also omit $z$ if $z = \la\ra$ and $\nu$ if $\nu = \zeroclocks$.
In contrast to \esg, we are only interested in finite traces, as \ac{MTL} is undecidable over infinite traces \cite{ouaknineDecidabilityMetricTemporal2005}.


Using the program transition semantics, we can now define the truth of a formula:
\begin{definition}[Truth of Formulas]\label{def:tesg-truth}
  Given a world $w \in \mathcal{W}$ and a formula $\alpha$, we define
  $w \models \alpha$ as $w,\langle\rangle, \vec{0} \models \alpha$, where for any $z \in
  \mathcal{Z}$ and clock valuations $\nu$:
  \begin{enumerate}
    \item $w, z, \nu \models F(n_1,\ldots,n_k)$ iff $w[F(n_1,\ldots,n_k),z] = 1$,
    \item $w, z, \nu \models (n_1 = n_2)$ iff $n_1$ and $n_2$ are identical,
    \item $w, z, \nu \models c \bowtie r$ iff $\nu(c) \bowtie r$,
    \item $w, z, \nu \models \alpha \wedge \beta$ iff $w, z, \nu \models \alpha$ and
      $w, z, \nu \models \beta$,
    \item $w, z, \nu \models \neg \alpha$ iff $w, z, \nu \not\models \alpha$,
    \item $w, z, \nu \models \forall x.\, \alpha$ iff $w, z, \nu \models \alpha^x_n$ for
      every standard name of the right sort,
    \item $w, z, \nu \models \square \alpha$ iff $w, z\cdot z', \nu' \models \alpha$
      for all $z' \in \mathcal{Z}$ and for all $\nu' \geq \nu$, and
    \item $w, z, \nu \models [\delta]\alpha$ iff for all 
      $(z', \nu') \in \|\delta\|^z_w$, $w, z \cdot z', \nu' \models \alpha$.
  \end{enumerate}
\end{definition}

Intuitively, $\square \alpha$ means that in every possible state, $\alpha$ is true, and $[\delta]\alpha$ means that \emph{after every execution} of $\delta$, $\alpha$ is true.




\subsection{Basic Action Theories}
\label{sec:BAT}

A \acf{BAT} defines the preconditions and effects of all actions of the domain, as well as the initial state:
\begin{definition}[Basic Action Theory]
  Given a finite set of fluent predicates $\mathcal{F}$ a finite set of clocks $\clockset$, and a set $\bat \subseteq \tesg$
  of sentences is called a \acf{BAT} over $(\mathcal{F},\clockset)$ iff
  $\bat = \bat_0 \cup \bat_\pre \cup \bat_g \cup \bat_\post$, where
  $\bat$ mentions only fluents in $\mathcal{F}$, clocks in $\clockset$, and
  \begin{enumerate}
    \item $\bat_0$ is any set of fluent sentences,
    \item $\bat_\pre$ consists of a single sentence of the form $\square \poss(a) \equiv \bigvee_o \{ \pi_o \}$, with one fluent formula $\pi_o$ with free variable%
    \footnote{Free variables are implicitly universal quantified from the outside. The modality $\Box$ has lower syntactic precedence than the connectives, and $[\cdot]$ has the highest priority.}
      $a$ for each action type $o$,
    \item $\bat_g$ is a set of sentences, one for each action $a$, of the form $\square \clockconstraints(a) \equivspace g_a$, where $g_a$ is a clock formula over $\clockset$, and \todo{We may want to allow multiple transitions}
    \item $\bat_\post$ is a set of sentences:
      \begin{itemize}
        \item one for each fluent predicate $F \in \mathcal{F}$, of the form $\square[a]F(\vec{x}) \equivspace
      \gamma_F$, where $\gamma_F$ is a fluent sentence, and
      \item one for each clock $c \in \clockset$, of the form $\square[a]\reset(c) \equivspace \gamma_c$, where $\gamma_c$ is a fluent sentence.
      \end{itemize}
  \end{enumerate}
\end{definition}

The set $\bat_0$ describes the initial state, $\bat_\pre$ defines the action preconditions, and
$\bat_\text{post}$ defines action effects by specifying for each fluent and clock of the domain whether the fluent
is true after doing some action $a$ and whether the respective clock is reset to zero.
The sentences in $\bat_{\text{g}}$ use clock formulas to describe the clock constraints of each action $a$.

We can now define \emph{programs}:

\begin{definition}[Program]
  A program is a pair $\Delta = (\bat, \delta)$ consisting of a \ac{BAT} \bat and a program expression $\delta$.
\end{definition}

We will later refer to the reachable subprograms of some program $\delta$:
\begin{definition}[Reachable Subprograms]
  Given a program $(\bat, \delta)$, 
  we define the \emph{reachable subprograms $\sub(\delta)$ of $\delta$}: 
\[
  \sub(\delta) = \{ \delta' \smid \exists w \models \bat, z \in \traces \text{ such that } \la\la\ra, \delta\ra \warrow^* \la z, \delta' \ra\}
\]
\end{definition}

\begin{example}[\acp{BAT}]\label{ex:bat}
We consider a system of a robot that can pick up (\emph{grasp}) an object $o$, 
with the following preconditions:
	\begin{align}
		\bat_\pre^{\mathit{hi}} =\{& \exists o,l.\ a = \sac{\mathit{grasp}(o,l)} \label{eq:startGrasp}\\
					 & \land\mathit{obj\_at}(o,l) \land \neg\mathit{grasping}(o)\land \neg\mathit{holding}(o),\nonumber\\
					 & \exists o,l.\ a = \eac{\mathit{grasp}(o,l)} \land\mathit{grasping}(o)\}.\label{eq:endGrasp}
	\end{align}
	To grasp an object $o$ at location $l$, the object needs to be at that location and the robot cannot already be grasping nor holding the object (\ref{eq:startGrasp}).
	Grasping can only end when the robot is currently grasping (\ref{eq:endGrasp}).
	Here we assume that the high-level \ac{BAT} does not specify any operations on clocks, i.e., $\bat_g = \emptyset$.
	The effects of actions on fluent predicates are specified by
	\begin{align}
		\bat_\post^{\mathit{hi}} =
					  & \square[a] \mathit{grasping}(o) \equivspace \exists l.\ a = \sac{\mathit{grasp}(o,l)}\label{eq:grasping}\\
					  & \lor\mathit{grasping}(o) \land a \neq \eac{\mathit{grasp}(o,l)},\nonumber\\
					  & \square[a] \mathit{holding}(o) \equivspace \exists l.\ a = \eac{\mathit{grasp}(o,l)}\label{eq:holding}\\
					  & \lor\mathit{holding}(o) \land \not\exists l',o'.\ a = \sac{\mathit{grasp}(o',l')},\nonumber\\
					  & \square[a] \mathit{obj\_at}(o,l) \equivspace \mathit{obj\_at}(o,l)\label{eq:objAt}\\
					  & \land a \neq \sac{\mathit{grasp}(o,l)}\}.\nonumber
	\end{align}
	The effects on the fluent predicates are detailed in \autoref{eq:grasping} to \autoref{eq:objAt},
  e.g., $\mathit{holding}(o)$ (\autoref{eq:holding}) is satisfied at the end of a grasping action of an object $o$, or if the robot is currently holding $o$ and does not start to grasp again.
	The initial situation, i.e, the initial satisfaction of the fluent predicates is described by
	\begin{align*}
		\bat_0^{\mathit{hi}} = \{ &\mathit{obj\_at}(o, l), \neg\mathit{holding}(o), \neg\mathit{grasping}(o) \}.
	\end{align*}
	The correct operation of the gripper requires a camera-module to be operational before starting to grasp for an object. These low-level operations are reflected by a secondary \ac{BAT} over a clock set $\clockset = \{c\}$ as follows:
	\begin{align*}
		\bat_\pre^{\mathit{lo}} = \{&a = \sac{\mathit{cam}()} \land \neg\mathit{cam\_on}(), \\
									 &a = \eac{\mathit{cam}()} \land \mathit{cam\_booting}()\}.
	\end{align*}
	Enabling the camera takes one time unit, which is outlined by the constraint
	\begin{align*}
		\bat_g^{\mi{lo}} = \{ \square g[\eac{\mathit{cam}()] \equivspace  c = 1} \}.
	\end{align*}
	The low-level \ac{BAT} specifies the following effects of actions:
	\begin{align*}
		\bat_\post^{\mathit{lo}} = \{ &\square[a] \mathit{cam\_on}() \equivspace a = \eac{\mathit{cam}()},\\
									   &\square[a] \mathit{cam\_booting}() \equivspace a = \sac{\mathit{cam}()}\\
									   &\lor \mathit{cam\_booting}() \land a \neq \eac{\mathit{cam}()},\nonumber\\
									   &\square[a] \mathit{reset(c)} \equivspace a = \sac{\mathit{cam}()} \}.\nonumber
	\end{align*}
	Initially, $\bat_0^{\mathit{lo}} = \{ \neg\mathit{cam\_on}(), \neg\mathit{cam\_booting}()\}$ holds for the low-level components.

\end{example}

\subsection{Finite-Domain \acp{BAT}}

As we allow quantification over the infinite set of standard names, a \ac{BAT} may generally describe infinite domains and thus also infinitely many fluents.
However, as we later want to specify \ac{MTL} constraints on the fluents entailed by some program configuration and \ac{MTL} requires a finite alphabet, we need to restrict the \ac{BAT} to a finite domain.
We do this by requiring a restriction on the quantifiers used in \bat:
\newcommand*{\primes}[1][\bat]{\ensuremath{\mathcal{P}_{#1}}\xspace}
\begin{definition}[Finite-domain \ac{BAT}]
  We call a \ac{BAT} $\bat$ a \emph{\ac{fd-BAT}} iff
  \begin{enumerate}
    \item each $\forall$ quantifier in $\bat$ occurs as $\forall x.\, \tau_i(x) \supset \phi(x)$,
      where $\tau_i$ is a rigid predicate, $i=o$ if $x$ is of sort object, and $i=a$ if $x$ is of sort action; \label{fd:quantifiers}
    \item $\bat_0$ contains axioms
      \begin{itemize}
      \item $\tau_o(x) \equiv (x = n_1 \vee x = n_2 \vee \ldots \vee x = n_k)$ and
      \item $\tau_a(a) \equiv (a = m_1 \vee a = m_2 \vee \ldots \vee a = m_l)$
      \end{itemize}
       where the $n_i$ and $m_j$ are object and action standard names,
       respectively. Also each $m_j$ may only mention object standard names $n_i$.
  \end{enumerate}
\end{definition}
We call a formula $\alpha$ that only mentions symbols and standard names from $\bat$ \emph{restricted to $\bat$}.
We denote the set of primitive formulas restricted to $\bat$ as \primes, the action standard names
mentioned in $\bat$ as $A_\bat$, and the clock standard names mentioned in $\bat$ as $C_\bat$.
We also write $\exists x\mathbf{:}i.\,\phi$ for $\exists x.\, \tau_i(x) \wedge \phi$ and $\forall x\mathbf{:}i.\,\phi$
for $\forall x.\, \tau_i(x) \supset \phi$.
Since an \ac{fd-BAT} essentially restricts the domain to be finite, quantifiers of type object can be understood as abbreviations, where
  $\exists x\mathbf{:}\tau_o. \phi \eqdef \bigvee_{i=1}^k \phi^x_{n_i}$,
  $\forall x\mathbf{:}\tau_o. \phi \eqdef \bigwedge_{i=1}^k \phi^x_{n_i}$,
and similarly for action-quantifiers.

We can now define an equivalence relation between worlds, where two worlds are equivalent if they initially satisfy the same fluents:
\newcommand*{\equivbat}{\ensuremath{\equiv_{\bat_0}}\xspace}
\begin{definition}[$\bat_0$-equivalent Worlds]
  We define the equivalence relation \equivbat of worlds wrt $\bat$ such that
  $w \equivbat w'$ iff $w[f, \la\ra] = w'[f, \la\ra]$ for all $f \in \primes$.
\end{definition}

We use $[w]$ to denote the equivalence class $[w] = \{ w' \in \worlds \mid w \equivbat w' \}$. 

As a \ac{fd-BAT} only refers to finitely many fluents, it also only has finitely many equivalence classes:
\begin{lemma}\label{thm:finite-w-equivalence-classes}
  \footnote{\ifthenelse{\boolean{proofappendix}}{Proofs are available in the appendix.}{Proofs are available in the supplementary material.}}
  For a \ac{fd-BAT} $\bat$, \equivbat has finitely many equivalence classes.
\end{lemma}

Furthermore, for each equivalence class, we only need to consider one world:
\begin{theorem}\label{thm:fd-bat-single-world}
  Let $\bat$ be a \ac{fd-BAT} and $w, w'$ be two worlds with $w \models \bat$, $w' \models \bat$, and $w \equivbat w'$.
  Then, for every formula $\alpha$ restricted to \bat:
  \[
    w \models \alpha \text{ iff } w' \models \alpha.
  \]
  \begin{fullproof}{\autoref{thm:fd-bat-single-world}}
    This is an adaption of Lemma 3 from \cite{lakemeyerSemanticCharacterizationUseful2011}.
    We first show that for any $w$, there exists a world $w_\bat$ that is the same as $w$, but satisfies $\bat_\pre \cup \bat_g \cup \bat_\post$.
    We define $w_\bat $ as the world that satisfies the following conditions:
    \begin{enumerate}
      \item For $F \not\in \fluents$ and for every $z \in \traces$, $w_\bat[F(\vec{n}), z] = w[F(\vec{x}), z]$
      \item For $F \in \fluents$, $w_\Sigma[F(\vec{n}), z]$ is defined inductively:
        \begin{enumerate}
          \item $w_\bat[F(\vec{n}), \la\ra ] = w[F(\vec{n}), \la\ra]$,
          \item $w_\bat[F(\vec{n}), z \cdot m] = 1$ iff $w_\Sigma, z \models (\gamma_F)^{a v_1 \cdots v_k}_{m n_1 \cdots n_k}$,
        \end{enumerate}
      \item $w_\bat[\poss(n), z] = 1$ iff $w_\bat, z \models (\pi_n)^a_n$.
    \end{enumerate}
    The argumentation is the same as in \cite{lakemeyerSemanticCharacterizationUseful2011}:
    $w_\bat$ clearly exists.
    The uniqueness follows from the fact that $\pi$ is a fluent formula and that for all fluents
    in $\fluents$, once their initial values are fixed, then the values after any number of actions are uniquely determined by $\bat_\text{post}$.
    Note that in particular, $\pi$ and $\bat_\text{post}$ may not mention clocks.
    Therefore, fluent values do not depend on time.

    Now, as each $w_\bat$ is unique, $w$ and $w'$ agree on the fluents in $\bat_0$, and $\alpha$ only mentions fluents from \bat, it follows that $w, z \models \alpha$ iff $w', z \models \alpha$.
  \end{fullproof}
\end{theorem}

Thus, for the sake of simplicity, for a given $\bat$, we assume in the following that we have a single world $w$ with $w \models \bat$.
To extend this to \acp{fd-BAT}, we can apply the proposed method for a world from each equivalence class.

\newcommand*{\actions}{\ensuremath{A}\xspace}
\newcommand*{\tactions}{\ensuremath{\mi{TA}}\xspace}

\section{\acf{MTL}}\label{sec:mtl}
While \autoref{sec:timed-esg} described how we can model \golog programs with \acp{BAT}, we now describe how we specify temporal constraints with \ac{MTL}.
\ac{MTL} \cite{koymansSpecifyingRealtimeProperties1990} extends \ac{LTL} with timing constraints on the \emph{Until} modality, therefore allowing temporal constraints with interval restrictions, e.g., $\fut{\leq 2} b$ to say that within the next two timesteps, a $b$ event most occur.  One commonly used semantics for
\ac{MTL} is a \emph{pointwise semantics}, in which formulas are interpreted over timed words.
We summarize \mtl{} and its pointwise semantics following the notation by \cite{ouaknineRecentResultsMetric2008}.
Timed words in \mtl are similar to timed traces in \tesg:
\begin{definition}[Timed Words]
	A timed word $\rho$\todo{We use rho for timed words and for remaining programs} over a finite set of atomic propositions $P$ is a finite or infinite sequence
  $\left(\sigma_0,\tau_0\right)\left(\sigma_1,\tau_1\right)\ldots$
  where $\sigma_i \subseteq P$ and $\tau_i \in \realpos$ such that the sequence $(\tau_i)$ is monotonically
  non-decreasing and non-Zeno.
  The set of timed words over $P$ is denoted as $\mi{TP}^*$.
\end{definition}

In contrast to the usual definition, we expect each symbol $\sigma_i$ to be a subset (rather than a single element) of the alphabet $P$.
We do this because we later want to define constraints over sets of fluents $F_i$ that are satisfied in some program state.

%

\mtl formulas are constructed as follows:
\begin{definition}[Formulas of \mtl{}]
  Given a set $P$ of atomic propositions, the formulas of \mtl{} are built as follows:
  \[
    \phi ::= p \smid \neg \phi \smid \phi \wedge \phi \smid \phi \until{I} \phi
  \]
\end{definition}

As an example, the formula $\mi{cam\_on} \until{[1, 2]} \mi{grasping}(o)$ says that the object $o$ must be grasped in the interval $[1, 2]$ and until then, the camera must be on.

We use the abbreviations
$\fut{I} \phi \eqdef (\top \until{I} \phi)$ (\emph{finally}) and
$\glob{I}\phi \eqdef \neg \fut{I} \neg \phi$
(\emph{globally}).


\begin{definition}[Pointwise Semantics of \mtl{}]\label{def:mtl-semantics}
  Given a timed word $\rho = \left(\sigma_1, \tau_1\right) \ldots$ over
  alphabet $P$ and an \mtl{} formula $\phi$, $\rho, i \models \phi$ is defined
  as follows:
  \begin{enumerate}
    \item $\rho, i \models p$ iff $p \in \sigma_i$,
    \item $\rho, i \models \neg \phi$ iff $\rho, i \not\models \phi$,
    \item $\rho, i \models \phi_1 \wedge \phi_2$ iff $\rho_i \models \phi_1$ and $\rho_i \models \phi_2$, and
    \item $\rho, i \models \phi_1 \until{I} \phi_2$ iff there exists $j$ such that
      \begin{enumerate}
        \item $i < j < |\rho|$,
        \item $\rho, j \models \phi_2$,
        \item $\tau_j - \tau_i \in I$,
        \item and $\rho, k \models \phi_1$ for all $k$ with $i < k < j$.
      \end{enumerate}
  \end{enumerate}
\end{definition}

For an \ac{MTL} formula $\phi$, we also write $\rho \models \phi$ for $\rho, 0 \models \phi$
and we define the language of $\phi$ as $\mathcal{L}(\phi) = \{ \rho \mid \rho \models \phi \}$.

Note that we use strict-until, i.e., we require that $i < j$ rather than $i \leq j$.
However, weak-until can be expressed with strict-until ($ \phi \textbf{U}_I^{\text{weak}} \psi \eqdef \psi \vee \phi \until{I} \psi)$, while strict-until cannot be expressed with weak-until \cite{henzingerItTimeRealtime1998}.

\subsection{Alternating Timed Automata}\label{sec:ata}

Alternating timed automata\acused{ATA}~(\acp{ATA}) \cite{ouaknineDecidabilityMetricTemporal2005} are a commonly used method to decide whether a timed word $\rho$ satisfies an \mtl formula $\phi$. 
We summarize the construction of an \ac{ATA} \ata which accepts a word $\rho$ iff the word satisfies formula $\phi$, i.e., iff $\rho \in \mathcal{L}(\phi)$.
We refer to \textcite{ouaknineDecidabilityMetricTemporal2005} for the full construction.

\begin{definition}
Let $L$ be a finite set of locations. The set of formulas $\Phi(L)$ is generated by the following grammar:
  \[
    \varphi ::= \top \smid \bot \smid \varphi_1 \wedge \varphi_2 \smid \varphi_1 \vee \varphi_2 \smid l \smid x \bowtie k \smid x.\varphi
  \]
where $k \in \mathbb{N}$, $\operatorname{\bowtie} \in \{ <, \leq, =, \geq, > \}$, and $l \in L$.
\end{definition}

\begin{definition}[\ac{ATA}]
	An \acf{ATA} is a tuple $\ata[] = \left(\ataalphabet, \atalocations, l_0, F, \delta\right)$, where
    \begin{itemize*}[label={}]
      \item $\ataalphabet$ is a finite alphabet,
      \item $\atalocations$ is a finite set of locations,
      \item $F \subseteq \atalocations$ is a set of accepting locations, and
      \item $\delta: \atalocations \times \ataalphabet \rightarrow \Phi(\atalocations)$ is the transition function.
    \end{itemize*}
\end{definition}

An \ac{ATA} has an implicit single clock $x$.
A state of \ata[] is a pair $(l, v)$, where $l \in \atalocations$ is the location and $v \in \realpos$ is a \emph{clock valuation} of the clock $x$.
We denote the set of all possible states with $\atastates$.
Given a set of states $M \subseteq\atastates$ and a clock valuation $v \in \realpos$, the truth of a formula $\varphi \in \Phi(\atalocations)$ is defined as follows:
\begin{enumerate*}[label=(\arabic*)]
  \item $(M, v) \models s$ iff $s \in M$,
  \item $(M, v) \models x \bowtie k$ iff $v \bowtie k$,
  \item $(M, v) \models x.\varphi$ iff $(M, 0) \models \varphi$.
\end{enumerate*}
The set of states $M$ is a \emph{minimal model of $\varphi$} if $(M, v) \models \varphi$ and there is no proper subset $N \subset M$ with $(N, v) \models \varphi$.
A \emph{configuration} of \ata[] is a finite set of states. The initial configuration is $\{(l_0, 0)\}$.
A configuration $\ataconf$ is accepting if for all $(l, u) \in \ataconf$, $l \in F$.

The language accepted by an \ac{ATA} is defined in terms of a transition system $\mathcal{T}_{\ata[]} = (2^\atastates, \rightsquigarrow, \rightarrow)$ over sets of configurations:
The time-labeled transition relation $\operatorname{\rightsquigarrow} \subseteq 2^\atastates \times \realpos \times 2^\atastates$ captures the progress of time in so-called \emph{flow steps}, where
$\ataconf \overset{t}{\rightsquigarrow} \ataconf'$ if $\ataconf' = \{ (s, v+t) \mid (s, v) \in \ataconf \}$.
The $\ataalphabet$-labeled transition relation $\operatorname{\rightarrow} \subseteq 2^\atastates \times \ataalphabet \times 2^\atastates$ describes \emph{edge steps}, instantaneous changes in the locations.
With $\ataconf = \{ (s_i, v_i) \}_{i \in I}$, the transition $\rightarrow$ is defined such that $\ataconf \overset{a}{\rightarrow} \ataconf'$ if $\ataconf' = \bigcup_{i \in I} M_i$, where $(M_i, v_i)$ is some minimal model of $\delta(s_i, a)$.
For a (finite) timed word $\rho = (\sigma_0, \tau_0), \ldots, (\sigma_k, \tau_k)$, the \emph{run} of \ata[] on $\rho$ is a sequence of alternating edge and flow steps:
\[
  \ataconf_0 \overset{\sigma_0}{\rightarrow} \ataconf_1 \overset{d_0}{\rightsquigarrow} \ataconf_2 \ldots \overset{\sigma_k}{\rightarrow} \ataconf_{2k} \overset{d_k}{\rightsquigarrow} \ataconf_{2k+1}
\]
A finite timed word $\sigma$ is accepted by \ata[] if there is a run of \ata[] over $\sigma$ leading to an accepting configuration.
We denote by $\mathcal{L}(\ata[])$ the set of timed words accepted by \ata[].

\begin{theorem}[\cite{ouaknineDecidabilityMetricTemporal2005}]
	Given an \ac{MTL} formula $\phi$, one can build an \ac{ATA} $\ata$ with $\mathcal{L}(\phi) = \mathcal{L}(\ata)$.
\end{theorem}
We omit the construction, but instead provide an example:
\begin{example}[ATA]
	Given the \ac{MTL} formula $\phibad = \top \until{\leq 1} \neg \camon \land \grasping$, \ata[\phibad] looks as follows:
  \begin{alignat*}{2}
    \ataalphabet &= 2^{\{\camon, \grasping \}} &\atalocations &= \{ \phibad \}
    \\
    l_0 &= \phibad &F &= \{\}
  \end{alignat*}
  \begin{alignat*}{2}
    \delta &= \{
           &(\phibad, \{ \}) &= \phibad,
           \\
           &&(\phibad, \{ \camon \}) &= \phibad,
           \\
           &&(\phibad, \{\grasping \}) &= x \leq 1 \vee \phibad,
      \\
           &&(\phibad, \{\camon, \grasping \}) &= \phibad
      \}
    \end{alignat*}
\end{example}


\section{The Control Problem}\label{sec:control-problem}

With the goal to control a \golog program against an \mtl specification
such that all execution traces satisfy the specification,
we define what we mean by a \emph{controller}, as well as what the \emph{control problem} is:

\newcommand*{\controllertraces}{\traces_{\controller}\xspace}
\begin{definition}[Controller]\label{def:controller}
Given a program $\Delta = (\delta, \bat)$,
a partition $A = A_E \dot\cup A_C$ of possible actions and an \ac{MTL} formula $\phi$,
a controller $\controller$ is a partial function
that maps a configuration to a set of successor configurations, i.e., $\controller(z, \nu, \delta) = \{ (z_i, \nu_i, \delta_i)_{i \in I} \}$ such that
\begin{enumerate}[align=left,label=(C\arabic*)]
	\item For each $z_i, \nu_i, \delta_i$: $\la z, \nu, \delta \ra \warrow \la z_i, \nu_i, \delta_i \ra$,\label{it:nonblocking}
	\item For each $a_e \in A_E$, if $\la z, \delta \ra \warrow \la z \cdot (a_e, t), \delta' \ra$, then $\la z \cdot (a_e, t), \delta' \ra \in \controller(z, \delta)$, and \label{it:nonrestricting}
	\item $\controller(z, \delta) = \{ \}$ implies $\la z, \delta\ra \in \mathcal{F}^w$.\label{it:terminating}
\end{enumerate}
\end{definition}

We demand that it is \emph{non-blocking}, i.e., it does not add deadlocks, and only maps to valid successors \ref{it:nonblocking}.
Second, we require the controller to be \emph{non-restrictive} \ref{it:nonrestricting}, i.e., it does not block the environment from executing its actions and third, we state that the controller may only terminate if a final configuration has been reached \ref{it:terminating}.

If we execute a controller by following its transitions iteratively and starting with the initial configuration $(\la\ra, \delta)$, we obtain a set of traces $\controllertraces$.
Formally, $\controllertraces$ is the smallest set such that
\begin{enumerate*}[label=(\arabic*)]
  \item for each $((a_1, t_1), \nu_1, \rho_1) \in \controller(\la\ra, \delta)$, there is a $z = \la (a_1, t_1), \ldots \ra \in \controllertraces$,
  \item for each $z \in \controllertraces$ 
and for each prefix $z_i$ of $z$,
if there are $\nu_{i}, \rho_{i}$ such that $(z_{i+1}, \nu_{i+1}, \rho_{i+1}) \in \controller(z_{i}, \nu_i, \rho_i)$, then
there is a $z^*$ such that $z_{i+1} \cdot z^* \in \controllertraces$.
\end{enumerate*}

In addition to action traces, we are also interested in the fluent traces (i.e., the satisfied fluents after each step):  
\begin{definition}[Fluent Trace]
  Given a world $w$ and a \ac{BAT} \bat, the \emph{fluent trace} $\fluenttrace(z)$ corresponding to a timed trace $z$ is the sequence
  $\fluenttrace(z) = \la (F_0, 0), (F_1, t_1) \ldots, (F_n, t_n)\ra$ where $F_i = \{ f \in \primes \mid w, z_i \models f \}$ and $t_i = \ztime(z_i)$.
\end{definition}

For each $z_i$, we also denote the last element $F_i$ of $\fluenttrace(z_i)$ with $F(z_i)$.
Furthermore, we denote the set of all fluent traces induced by the controller as $\fluenttraces_\controller = \{ \fluenttrace(z) \mid z \in \controllertraces \}$.
We can now define the control problem:

\begin{definition}[Control Problem]\label{def:control-problem}
 Given a program $\Delta = (\delta, \bat)$ and an \ac{MTL} formula $\phi$, the control problem is to determine a controller $\controller$ such that
 for each $\fluenttrace \in \fluenttraces_{\controller}$: $\fluenttrace \models \phi$.
\end{definition}

Intuitively, the problem is to determine a controller for the program $\Delta$ such that each execution satisfies the specification \fluenttrace, which is an \mtl formula over the fluents of $\bat$.

\section{Approach}\label{sec:approach}

The goal of this work is to synthesize a controller which orchestrates actions of a high-level program and a second, low-level program in such a way that a task is successfully executed while complying with a given specification.
In our scenario, the idea is, among safety requirements, to express platform requirements as constraints on the high-level program, which can be satisfied by correct concurrent execution of high- and low-level actions.

Our approach consists of the following steps:
\begin{enumerate*}[label=(\arabic*)]
  \item we construct an \ac{ATA} \ata from the specification,
  \item we define a labeled transition system \symlts that symbolically executes the \golog program,
  \item we combine \ata and \symlts to obtain a synchronous product \lts (which may have uncountably many states),
  \item we regionalize \lts by applying regionalization to the clock values to obtain \disq with countably many states,
  \item we apply a powerset construction to obtain a deterministic \detdisq, and
  \item we define a timed game on \detdisq that results in a decidable procedure to solve the synthesis problem.
\end{enumerate*}

To obtain a transition system of a program $\Delta$ we need to introduce several notions.
First of all, to describe sets of executions symbolically, we define symbolic traces:

\begin{definition}[Symbolic Alphabet and Symbolic Traces]
  Given a program $\Delta = (\delta, \bat)$, the \emph{symbolic alphabet} of $\Delta$ is the set $\symalphabet_{\Delta} = A_\bat \times \clockset_\bat \times 2^{\clockset_\bat}$. 
  A symbolic trace of $\Delta$ is a sequence $\symtrace = (a_1, g_1, Y_1)  \ldots \in \symalphabet_\Delta^*$.
\end{definition}
We denote the set of all symbolic traces of $\Delta$ with $\symtraces$.


\begin{definition}[Induced Timed Trace]\label{def:induced-time-trace}
  A symbolic trace $\symtrace = (a_1, g_1, Y_1) (a_2, g_2, Y_2) \ldots \in \symtraces$
  induces a set of timed traces $\tw(\symtrace)$ over $A_\bat \times \realpos$, where
  $\trace \in \tw(\symtrace)$ iff $|\trace| = |\symtrace|$, $\trace = (a_1, t_1)(a_2, t_2)\ldots$ and there is a sequence of valuations $\nu_0, \nu_1, \ldots$ (compatible with $g_1, \ldots$ and $Y_1, \ldots$) and programs $\rho_i \in \sub(\delta)$ such that
  $\la \la\ra, \nu_0, \delta\ra \warrow \la \la (a_1, t_1) \ra, \nu_1, \rho_1 \ra \warrow \ldots \warrow \la z, \nu_n, \rho_n \ra$.
\end{definition}

Symbolic traces are sufficient to describe the execution of a program, as all timed traces induced by a symbolic trace end in the same observable trace:

\begin{theorem}\label{thm:timed-trace-equiv}
  Let $\symtrace $ be a symbolic trace of a program $\Delta$, let $w \models \bat$, and $\trace, \trace' \in \tw(s)$.
  Then for every situation formula $\alpha$: $w, \trace \models \alpha$ iff $w, \trace' \models \alpha$.
  \begin{fullproof}{\autoref{thm:timed-trace-equiv}}
    First, $|z| = |z'| = n$ by definition of $\tw$.
    We show the statement by induction over $n$.
    \\
    \textbf{Base case.}
    Let $z = \la\ra$. Then also $z' = \la\ra$, and the statement follows.
    \\
    \textbf{Induction step.}
    Let $z = z_1 \cdot (a, t)$ and $z' = z_1' \cdot (a', t')$.
    By induction, $w, z \models \alpha$ iff $w, z' \models \alpha$.
    By \autoref{def:induced-time-trace}, $a = a'$.
    As the precondition and successor state axioms may not mention clock constraints, $w, z_1 \models [a] \alpha$ iff $w, z_1' \models [a] \alpha$.
    Thus, the statement follows.
  \end{fullproof}
\end{theorem}


\newcommand*{\wsigmaf}[1][F]{\ensuremath{w_{\bat}^{#1}}\xspace}
%

We can now define the symbolic execution of a program:
\begin{definition}[Symbolic Program Execution]
  Given a program $\Delta = (\delta, \bat)$, the \emph{symbolic execution} of $\Delta$ is a labeled transition system $\symlts = (\symstates, \symstate_0, \symtrans[]{})$ defined as follows:
  \begin{itemize}
    \item $\symstates = \symtraces \times \clockvaluations_C \times \sub(\delta)$,
    \item $\symstate_0 = (\la\ra, \zeroclocks, \delta)$,
    \item $(\symtrace, \nu, \rho) \symtrans{(a, g, Y)} (\symtrace', \nu', \rho')$ if there is $\trace \in \tw(\symtrace)$ and $\trace' = \trace \cdot (a, \ztime(\trace) + t)$ such that
      \begin{enumerate}
        \item $\symtrace' = \symtrace \cdot (a, g, Y)$,
        \item $\la \trace, \nu, \rho \ra \warrow \la \trace', \nu', \rho'\ra$,
        \item $c \in Y$ iff $w, \trace' \models \reset(c)$, and
        \item $g = g_a$, i.e., the guard in $\bat$ wrt $a$.
      \end{enumerate}
  \end{itemize}
\end{definition}

\begin{example}
  The \symlts of the program $\Delta = (\bat^{\text{lo}}, \delta)$ with
  \[
    \delta = (\sac{\acam}; \eac{\acam};) \mid \sac{\grasp}
  \]
  allows the following transitions:
  \[
    \symstate_0 \symtrans[0.5]{\sac{\acam}, \top, \{ \ccam \}} \symstate_1
    \quad
    \symstate_1 \symtrans[1]{\eac{\acam}, \ccam = 1, \{ \}} \symstate_2
  \]
  (and more), where
  \begin{align*}
    \symstate_0 &= (\la\ra, \zeroclocks, \delta)
    \\
    \symstate_1 &= (\la(\sac{\acam}, \top, c_{\mi{cam}})\ra, \{ c_{\mi{cam}} : 0 \},
    \\
                & \qquad \eac{\acam} \mid \sac{\grasp})
    \\
    \symstate_2 &= (\la(\sac{\acam}, \top, c_{\mi{cam}}), (\eac{\mi{cam}()}, c_{\mi{cam}} = 1, \emptyset)\ra,
    \\
                &\qquad \{ c_{\mi{cam}} : 1 \}, \sac{\grasp})
  \end{align*}

\end{example}

For $\symstate = (\symtrace, \nu, \rho)$, we also write $\tw(\symstate)$ to mean $\tw(\symtrace)$.
For a path $p  = (\la\ra, \zeroclocks, \delta) \symtrans{(a_1, g_1, Y_1)} (\symtrace_1, \nu_1, \rho_1)  \symtrans{(a_2, g_2, Y_2)} \ldots \symtrans{(a_n, g_n, Y_n)} (\symtrace_n, \nu_n, \rho_n)$ of $\symlts$, we write $\traces(p) =\tw(\symtrace_n) $ for the timed traces induced by $p$.
We say $p$ ends in a final configuration if there is a $\trace \in \tw(\symtrace_n)$ such that $\la \trace, \nu_n,  \rho_n \ra \in \mathcal{F}^w$.
We write $\traces(\symlts)$ for the set of induced traces of all paths in $\symlts$ ending in a final configuration.

\begin{lemma}\label{lma:all-z-final}
  Let $p$ be a path $p  = (\la\ra, \zeroclocks, \delta) \symtrans{(a_1, g_1, Y_1)} (\symtrace_1, \nu_1, \rho_1)  \symtrans{(a_2, g_2, Y_2)} \ldots \symtrans{(a_n, g_n, Y_n)} (\symtrace_n, \nu_n, \rho_n)$ of \symlts ending in a final configuration.
  Then for every $\trace \in \traces(p)$, $\la \trace, \nu_n, \rho_n \ra \in \mathcal{F}^w$.
  \begin{fullproof}{\autoref{lma:all-z-final}}
    As $p$ is final, there is a $z_f \in \traces(p)$ such that $\la z_f, \nu_n, \rho_n \ra \in \mathcal{F}^w$.
    Also, by definition of $\traces(p)$, $z \in \traces(p)$ iff $z \in \tw(\symtrace_n)$.
    Thus, by \autoref{thm:timed-trace-equiv}, for every $z \in \traces(p)$ and every situation formula $\alpha$: $w, z \models \alpha$ iff $w, z_f \models \alpha$.
    Thus, $\la z, \nu_n, \rho_n \ra \in \mathcal{F}^w$.
  \end{fullproof}
\end{lemma}

Next, the following theorem establishes the validity of a symbolic execution with respect to explicit program traces.

\begin{theorem}\label{thm:symbolic-execution}
  $\trace \in \traces(\symlts)$ iff $\trace \in \|\delta\|_w$.
  \begin{fullproof}{\autoref{thm:symbolic-execution}}
    \textbf{$\Rightarrow$:}
    Let $\trace \in \traces(\lts)$. Then there is a path $p  = (\la\ra, \zeroclocks, \delta) \symtrans{(a_1, g_1, Y_1)} (\symtrace_1, \nu_1, \rho_1)  \symtrans{(a_2, g_2, Y_2)} \ldots \symtrans{(a_n, g_n, Y_n)} (\symtrace_n, \nu_n, \rho_n)$
    such that for some $\trace_f \in \tw(\symtrace_n)$, $\la \trace_f, \nu_n, \rho_n \ra \in \mathcal{F}^w$.
    By definition of \symtrans[]{} of \symlts, $\la \la\ra, \zeroclocks, \delta\ra \warrow^* \la \trace, \nu_n, \rho_n\ra$.
    Furthermore, by \autoref{lma:all-z-final}, $\la \trace, \nu_n, \rho_n\ra \in \mathcal{F}^w$.
    Therefore, $\trace \in \|\delta\|_w$.
    \\
    \textbf{$\Leftarrow$:}
    Let $\trace \in \|\delta\|_w$. Then, by definition of $\|\cdot\|_w$, $\la \la\ra, \zeroclocks, \delta \ra \warrow^* \la \trace, \nu_n, \rho_n\ra$ and $\la \trace, \nu_n, \rho_n \ra \in \mathcal{F}^w$.
    Now, we show by induction over the length $i$ of $\trace$ that $\la \la\ra, \zeroclocks, \delta \ra \warrow^* \la \trace, \nu_i, \rho_i\ra$ implies that there is a path
    $p  = (\la\ra, \zeroclocks, \delta) \symtrans[t_1]{(a_1, g_1, Y_1)} (\symtrace_1, \nu_1, \rho_1)  \symtrans[t_2]{(a_2, g_2, Y_2)} \ldots \symtrans[t_i]{(a_i, g_i, Y_i)} (\symtrace_i, \nu_i, \rho_i)$ with $z \in \tw(s_i)$.
    \\
    \textbf{Base case.}
    With $i = 0$, it follows that $z = \la\ra$ and $|p| = 0$. Clearly, $\trace \in \tw(\la\ra)$.
    \\
    \textbf{Induction step.}
    Let $|z| = |p| = i$ and $z' = z \cdot (a, t)$.
    By \subdefref{def:trans}{def:trans:action}, $\la z, \nu_i, \rho_i \ra \warrow \la z', \nu_{i+1}, \rho_{i+1} \ra$ and there is a $d \geq 0$ such that
    \begin{enumerate}
      \item $t = \ztime(z) + d$,
      \item $w, z \models \poss(a)$,
      \item $w, z, \nu_i + d \models \clockconstraints(a)$,
      \item $\nu_{i+1} = 0$ if $w, z' \models \reset(c)$ and $\nu_i + d$ otherwise.
    \end{enumerate}
    Thus, $z' \in \tw(s \cdot (a, g_a, Y))$, where $c \in Y$ iff $w, z' \models \reset(c)$.
    Then, by definition of \symtrans[]{} of \symlts, $(s, \nu_i, \rho_i) \symtrans[d]{a, g_a, Y} (s', \nu_{i+1}, \rho_{i+1})$.
  \end{fullproof}
\end{theorem}

With this symbolic abstraction of the input program, we can now create the product automaton of the symbolic execution $\symlts$ and the \ac{ATA} \ata which is used to track the satisfaction of the input specification.

\begin{definition}[Synchronous Product]
Given a symbolic execution $\symlts = (\symstates, \symstate_0, \symtrans[]{})$ and an \ac{ATA} \ata, the synchronous product $\lts = (\syncstates, \syncstate_0, \rightarrow)$ is a labeled state transition system, defined as follows:
\begin{itemize}
  \item $\syncstates = \symstates \times \ataconfs$,
  \item $\syncstate_0 = (\symstate_0, \ataconf_0)$,
  \item $(\symstate, \ataconf) \synctrans{(a, g, Y)} (\symstate', \ataconf')$ if $\symstate \symtrans{(a, g, Y)} \symstate'$ and
    $\ataconf \overset{t}{\rightsquigarrow} \ataconf^* \overset{F}{\rightarrow} \ataconf'$ with $\trace \in \tw(e)$ and $f \in F$ iff $w[f, \trace] = 1$.
\end{itemize}
\end{definition}

\begin{example}\label{ex:sync-product}
	The synchronous product \lts of \symlts and \ata[\phi_{bad}] (where $\phibad = \top \until{\leq 1} \neg c \land g$) looks as follows:\todo{Wir sollten noch erwähnen, dass die t's repräsentanten sind?}
  \begin{align*}
    \syncstate_0 &= (e_0, G_0) = ((\la\ra, \zeroclocks, \delta), \{ (\phibad, 0) \})
    \\
    (e_0, G_0) &\symtrans[0.5]{\sac{\acam}, \top, \{ \ccam \}} (e_1, \{ (\phibad, 0.5) \})
    \\
    (e_1, \{ (\phibad, 0) \}) &\symtrans[1]{\eac{\acam}, c = 1, \emptyset} (e_2, \{ (\phibad, 1.5) \} )
    \\
    (e_0, G_0) &\symtrans[0.5]{\sac{\grasp}, \top, \emptyset} (e_1, \emptyset)
  \end{align*}
  The last transition ends in a configuration with an empty \ac{ATA} configuration, as \grasp causes \grasping to be true and therefore satisfies the specification of bad behavior.
\end{example}

For a symbolic trace $\symtrace$, we write $\fluenttrace(\symtrace)$ for the fluent trace $\fluenttrace(\trace)$ and $F(\symtrace)$ for the set of satisfied fluents $F(\trace)$, where $\trace \in \tw(\symtrace)$.
Note that this is well-defined, as by \autoref{thm:timed-trace-equiv}, $\fluenttrace(\trace) = \fluenttrace(\trace')$ for every $\trace, \trace' \in \tw(\symtrace)$.

\begin{theorem}\label{thm:traces-of-syncproduct}
  Given a synchronous product \lts of \symlts and \ata, and a timed trace $\trace$. Then:
  \begin{enumerate}
    \item If $\trace \in \traces(\lts)$, then there is also a run of $\ata$ on $F(\trace)$.
    \item $\trace \in \traces(\lts)$ iff $\trace \in \|\delta\|_w$.
  \end{enumerate}
  \begin{fullproof}{\autoref{thm:traces-of-syncproduct}}
    ~
    \begin{enumerate}
      \item Follows directly from the fact that $\ata$ is complete.
      \item Follows by \autoref{thm:symbolic-execution} and because $\ata$ is complete. \qedhere
    \end{enumerate}
  \end{fullproof}
\end{theorem}

\newcommand*{\regions}[1][K]{\ensuremath{\text{REG}_{#1}}\xspace}
\subsection{Regionalization}
We regionalize the states $S$ of \lts to obtain a canonical representation and a finite abstraction for states in $\lts$, inspired by~\cite{ouaknineDecidabilityMetricTemporal2005,bouyerControllerSynthesisMTL2006,alurTheoryTimedAutomata1994}.
Let $K$ be the greatest constant mentioned in $\phi$.
Then \regions is a finite set of regions defined for each $0 \leq i \leq K$ as follows: $r_{2i} = \{i\}$, $r_{2i+1} = (i, i+1)$ and the last region collecting all values larger than $K$ as $r_{2K+1} = (K, \infty)$.
For $u \in \realpos$, $\reg(u)$ denotes the region in \regions containing $u$.
For each \lts-state $((\symtrace, \nu, \rho), \ataconf)$, we compute its canonical representation.
Let $\Lambda = 2^{(\clockset \cup L) \times \regions}$ be the alphabet over regionalized clocks (tuples of name and region) used to represent a regionalized state as follows:

First, we represent the clock values in $\nu$ as $G_\nu = \{ (c, \nu(c)) \mid c \in \clockset \}$.
We partition $G_\nu \cup \ataconf$ into a sequence of subsets $G_1, \ldots, G_n$ such that for every $1 \leq i \leq j \leq n$, for every pair $(l_i, c_i) \in G_i$ and every pair $(l_j, c_j) \in G_j$, the following holds: $i \leq j$ iff $\fract(c_i) \leq \fract(c_j)$.
For each $G_i$, let $\abs(G_i) = \{ (l, \reg(c)) \mid (l, c) \in G_i \} \in \Lambda$.
Then, the canonical representation $H(\nu, \ataconf) \in \Lambda^*$ of $\nu$ and $\ataconf$ is defined as the sequence $H(\nu, \ataconf) = (\abs(G_1), \ldots, \abs(G_n))$.
We say that two \lts-configurations $\ltsstate = (\symtrace, \nu, \rho, \ataconf)$ and $\ltsstate' = (\symtrace', \nu', \rho', \ataconf')$ are equivalent, written $\ltsstate \sim \ltsstate'$ if $(\symtrace, \rho, H(\nu, \ataconf)) = (\symtrace', \rho', H(\nu', \ataconf'))$.



\begin{proposition}[\cite{bouyerControllerSynthesisMTL2006}]\label{thm:bisimulation}
  The relation $\sim$ is a bisimulation over $\lts$, i.e., $\ltsstate_1 \sim \ltsstate_1'$ and $\ltsstate_1 \trans{a, g, Y} \ltsstate_2$ implies $\ltsstate_1' \trans{a, g, Y} \ltsstate_2'$ for some $\ltsstate_2'$ with $\ltsstate_2 \sim \ltsstate_2'$.
\end{proposition}

Using this proposition, we can define a regionalized version of the synchronous product as follows:

\begin{definition}[Regionalized Synchronous Product]
  Given a synchronous product $\lts = (\syncstates, \syncstate_0, \synctrans{})$.
  The \emph{discrete quotient} \disq of $\lts$ is a \ac{STS} $\disq = (\disqstates, \disqstate_0, \regtrans{})$ with
  \begin{itemize}
    \item $\disqstates = \{ (\symtrace, \rho, H(\nu, \ataconf)) \mid ((\symtrace, \nu, \rho), \ataconf) \in S \}$,
    \item $\disqstate_0 = (\la\ra, \delta, H(\zeroclocks, \ataconf_0))$, and
    \item $(\symtrace, \rho, \regconf) \regtrans{a, g, Y} (\symtrace', \rho', \regconf')$ iff there exists $(\nu, \ataconf) \in H^{-1}(\regconf)$ such that such that $((\symtrace, \nu, \rho), \ataconf) \trans{a, g, Y} ((\symtrace', \nu', \rho'), \ataconf')$ and $\regconf' = H(\nu', \ataconf')$.
  \end{itemize}
\end{definition}

\begin{example}
  The regionalization of \lts from \autoref{ex:sync-product} with $K=1$ looks as follows:
  \begin{align*}
    \disqstate_0 &\regtrans{\sac{\acam}, \top, \{ \ccam \}} \disqstate_1
    \\
    \disqstate_1 &\regtrans{\eac{\acam}, c = 1, \emptyset} \disqstate_2
    \\
    \disqstate_0 &\regtrans{\sac{\grasp}, \top, \emptyset} \disqstate_3
  \end{align*}
  where
  \begin{align*}
    \disqstate_0 &= (\la\ra, \delta, (\{ (\ccam, 0), (\phibad, 0) \}))
    \\
    \disqstate_1 &= (\la(\sac{\acam}, \top, c_{\mi{cam}})\ra, \eac{\acam} \mid \sac{\grasp},
                 \\
                 &\qquad (\{ (\ccam, 0), (\phibad, 1) \}))
    \\
    \disqstate_2 &= (\la\ra, \sac{\grasp}, (\{ (\ccam, 2), (\phibad, 3) \}))
    \\
    \disqstate_3 &= (\la\ra, \eac{\acam}, (\{ (\ccam, 1) \}))
  \end{align*}
\end{example}
\begin{theorem}\label{thm:traces-of-disq}
  ~
  \begin{enumerate}
    \item If $\trace \in \traces(\disq)$, then there is also a run of $\ata$ on $F(\trace)$.
    \item $\trace \in \traces(\disq)$ iff $\trace \in \traces(\lts)$.
  \end{enumerate}
  \begin{fullproof}{\autoref{thm:traces-of-disq}}
    ~
    \begin{enumerate}
      \item Follows directly from the fact that $\ata$ is complete.
      \item Follows by the definition of $\regtrans{}$ and \autoref{thm:bisimulation}. \qedhere
    \end{enumerate}
  \end{fullproof}
\end{theorem}

With the definition of $\disq$ we have obtained a finite abstraction of the original product automaton of the input program and the specification.
However, the abstraction $\disq$ allows several successors for the same symbolic action $a$, as in $\disq$ states are  also distinguished based on the configurations of the \ac{ATA}.
To test whether a final configuration reachable via a symbolic trace is safe requires to check all possible successors for all symbolic actions on this trace for safety.
To overcome this, analogous to~\cite{bouyerControllerSynthesisMTL2006}, in the next step we make $\disq$ symbol-deterministic.

\begin{definition}[Deterministic Discrete Quotient]\label{def:detdiscquotient}
  Given a discrete quotient $\disq = (\disqstates, \disqstate_0, \regtrans{})$, the deterministic version $\detdisq = (\detstates, \detstate_0, \detregtrans{})$ of \disq is defined as follows:
  \begin{itemize}
    \item $\detstates = \symtraces \times \sub(\delta) \times 2^{\Lambda^*}$,
    \item $\detstate_0 = (\la\ra, \delta, \{ H(w_0) \})$, and
    \item $(\symtrace, \rho, \hset) \detregtrans{a, g, Y} (\symtrace', \rho', \hset')$ iff
      $\hset'  = \{ \regconf' \mid \exists \regconf \in \hset \text{ with } (\symtrace, \rho, \regconf) \regtrans{a, g, Y} (\symtrace', \rho', \regconf') \}$.
  \end{itemize}
\end{definition}

\begin{theorem}\label{thm:traces-of-det-disq}
  ~
  \begin{enumerate}
    \item If $\trace \in \traces(\detdisq)$, then there is also a run of $\ata$ on $F(\trace)$.
    \item $\trace \in \traces(\detdisq)$ iff $\trace \in \traces(\disq)$.
  \end{enumerate}
  \begin{fullproof}{\autoref{thm:traces-of-det-disq}} ~
	  \begin{enumerate}
		  \item Follows from the completeness of \ata.
		  \item Follows from the definition of $\regtrans{}$ of \disq and \autoref{thm:bisimulation}: the transition-relation of $\detdisq$ combines successor-states of a symbolic action $a,g,Y$, which per definition agree on the same $\symtrace,\rho,\nu$ and only differ in the \ac{ATA}-configuration. \qedhere
	  \end{enumerate}
  \end{fullproof}
\end{theorem}

\subsection{Timed Games}

We use a variant of \emph{downward closed games}~\cite{abdullaDecidingMonotonicGames2003,bouyerControllerSynthesisMTL2006} for the synthesis of a controller, where a controller exists if there is a safe strategy, i.e., a trace in the \golog program that leads to an accepting state while the specification is satisfied.
We construct a timed game over $\detdisq$, whose states allow to determine safety with respect to the specification, which enables us to formally describe a \emph{winning strategy} for the timed game.
Formally, a timed game is defined as follows:

\begin{definition}[Timed Golog Game]
  A \emph{timed \golog game} is a pair $\mathbb{G} = (\Delta, \mathcal{L})$, where $\Delta$ is a \golog program and $\mathcal{L} \subseteq T\primes^*$ is a timed language over finite words.
\end{definition}
  A \emph{validity function} over $A_\bat$ is a function $\val \colon 2^{\mi{A}_\bat} \rightarrow 2^{2^{\mi{A}_\bat}}$ such that for every set of timed actions $U \subseteq \mi{A}_\bat$ is mapped to a non-empty family of subsets of $U$.
  A \emph{strategy} in $\Delta$ respecting $\val$ is a mapping
  that maps each program state to a set of actions such that each of those actions again results in successor states that are mapped by the strategy. Formally, it is a mapping
  $f \colon D \subseteq \suc^*_w(\la\ra, \zeroclocks, \delta) \rightarrow 2^{\mi{A}_\bat}$ such that $(\la\ra, \zeroclocks, \delta) \in D$ and for all $s = (\trace, \nu, \rho) \in D$,
  $f(s) \in \val(\{ a \mid (\trace \cdot (a, t_a), \nu', \rho') \in \suc(s) \})$,
  and for all $b \in f(s)$ and every $s'$ with $s' = (\trace \cdot (b, t_b), \nu', \rho') \in \suc(\trace, \nu, \rho)$: $s' \in D$.
  The set of plays of $f$, denoted by $\plays(f)$, is the set of traces of $\delta$ that are consistent with the strategy $f$.
  Formally, $\trace \in \plays(f)$ iff for every prefix $\trace' \cdot b$ of $\trace$, $b \in f(\trace')$.

  Let $\val$ be a validity function over $\mi{A}_\bat$. A strategy respecting $\val$ in the timed game $\mathbb{G} = (\Delta, \mathcal{L})$ is a strategy in $\Delta$ respecting $\val$.
  A strategy $f$ is winning with respect to undesired behavior iff $(\plays(f) \cap \|\delta\|_w) \cap \mathcal{L} = \emptyset$.
Intuitively, a state in $\detdisq$ is bad, if it is final and contains a bad \ac{ATA}-configuration.
Formally, a \lts-state $a = ((\symtrace,\nu,\rho), \ataconf)$ is \emph{bad} if there exists a $\trace\in\tw(\symtrace)$ with $\la \trace, \nu, \rho \ra \in \mathcal{F}^w$ and $\ataconf$ is accepting.
A state $(\symtrace, \rho, h)$ of \disq is \emph{bad} if there exists $(\nu, \ataconf) \in H^{-1}(h)$ and $((\symtrace,\nu,\rho),\ataconf)$ is bad.
A state $(\symtrace,\rho,\mathcal{C})$ of $\detdisq$ is \emph{bad} if there is an $h\in\mathcal{C}$  such that $(\symtrace,\rho,h)$ is bad.
A strategy $f$ in $\detdisq$ is \emph{safe} iff for every finite play $\trace$ of $f$, $\trace$ does not end in a bad state of $\detdisq$.
\begin{theorem}\label{thm:winning-strategy}
  There is a winning strategy in $\mathbb{G}$ with respect to \emph{undesired behavior} iff there is a safe strategy in $\detdisq$.
  \begin{fullproof}{\autoref{thm:winning-strategy}}
    By \autoref{thm:traces-of-syncproduct}, \autoref{thm:traces-of-disq}, and \autoref{thm:traces-of-det-disq}, $\trace \in \traces(\detdisq)$ iff $\trace \in \|\delta\|_w$ for some $w \models \bat$.
    Therefore, for every strategy $f$, $f$ is a strategy in $\mathbb{G}$ iff $f$ is a strategy in $\detdisq$.
    If $f$ is a winning strategy in $\mathbb{G}$ with respect to undesired behavior, then we show that $f$ is safe for $\detdisq$:
    Suppose that $\trace$ is a bad play in $\detdisq$.
    Then, by definition of $\detdisq$ and by \autoref{thm:bisimulation}, there would be a path in $\lts$ from the initial state to a bad state whose trace is $\trace$. By construction, this implies $\trace \in \|\delta\|_w$ for some $w \models \bat$ and $\trace \in \mathcal{L}(\phi)$.
    Contradiction to $f$ being a winning strategy in $\mathbb{G}$.

    Similarly, if $f$ is safe for $\detdisq$, then we show that $f$ is a winning strategy in $\mathbb{G}$ with respect to undesired behavior.
    Suppose $f$ is not a winning strategy in $\mathbb{G}$.
    Then, there is a play $\trace \in \plays(f)$ with $\trace \in \|\delta\|_w$ for some $w \models \bat$ and $\trace \in \mathcal{L}(\phi)$.
    By definition of $\detdisq$ and by \autoref{thm:bisimulation}, $\trace$ is a bad play in $\detdisq$.
    Contradiction to $f$ being safe for $\detdisq$.
  \end{fullproof}
\end{theorem}

\subsection{Decidability}

With the definition of a timed \golog game from the previous section, we can design an algorithm, that synthesizes a controller for $\Delta$ with respect to $\phi$.
In the following, we show that the underlying problem of finding a winning strategy in our game is decidable.

Intuitively, the idea is to show that the search over states of $\detdisq$ terminates.
We define a reflexive and transitive relation $\leq$ on states $s_i$ of $\detdisq$, which allows to state that if $s_1 \leq s_2$ holds and we know that for $s_1$ we cannot find a solution, then we also cannot find a solution for state $s_2$.
If for every infinite sequence of states, there exists at least one pair of indices $i,j, i < j$ such that $s_i \leq s_j$, we can safely terminate the search when reaching $s_j$.
In case $\leq$ fulfills the this property, $(\detstates,\leq)$ is called a \emph{\wqo}.

To be able to relate two states $(s,\rho,\hset)$ and $(s',\rho',\hset')$ of $\detdisq$, we need to be able to relate the sets of regionalized configurations $\hset, \hset'$, which can be achieved via an induced \emph{monotone domination order}.
\begin{definition}[Monotone Domination Order]
  Given a \qo $(S, \leq)$, the \emph{monotone domination order} is the \qo $(S^*, \leq^*)$ over the set $S^*$ of finite words over $S$ such that $x_1, \ldots, x_m \leq^* y_1, \ldots, y_n$ iff there is a strictly monotone injection $h: \{ 1, \ldots, m \} \rightarrow \{ 1, \ldots, n \}$ such that $x_i \leq y_{h(i)}$ for all $1 \leq i \leq m$.
\end{definition}

\begin{example}[Monotone Domination]
	Consider the two finite sequences of sets of natural numbers $s_1 = ( \{1\}, \{2\})$ and $s_2 = (\{1\}, \{1,2\}, \{3\})$. Using the monotone domination order $(2^{\mathbb{N}*}, \preccurlyeq)$ induced by the \qo $(2^\mathbb{N}, \subseteq)$, we can see that $s_1 \preccurlyeq s_2$ since $\{1\} \subseteq \{1\}$ and $\{2\} \subseteq \{1,2\}$.

\end{example}

Here, we will use the \qo $(\Lambda, \subseteq)$ to induce a monotonic domination order for $\hset$.

We use results by~\cite{abdullaTimedPetriNets2001} which relate different orderings to state that the induced monotonic domination order here is a \wqo.

\begin{proposition}[\cite{abdullaTimedPetriNets2001}]\label{prop:bqos}
  \begin{enumerate*}[label=(\arabic*)]
    \item Each \bqo is a \wqo.
    \item If $S$ is finite, $(2^S \subseteq)$ is a \bqo.
    \item If $(S, \leq)$ is a \bqo, then $(S^*, \leq^*)$ is a \bqo.
    \item If $(S, \leq)$ is a \bqo, then $(2^S, \sqsubseteq)$ is a \bqo.
  \end{enumerate*}
\end{proposition}

It follows that while in general a monotonic domination order is a \qo, here it is a \wqo:

\newcommand*{\lpowleq}{\ensuremath{\sqsubseteq}}
\begin{lemma}\label{lma:lambda-star-bqo}
  ~
  \begin{itemize}
    \item The monotone domination order $(\Lambda^*, \preccurlyeq)$ induced by the \qo $(\Lambda, \subseteq)$ is a \bqo.
    \item The powerset ordering $(2^{\Lambda^*}, \lpowleq)$ induced by $(\preccurlyeq, \Lambda^* )$ is a \bqo.
  \end{itemize}
  \begin{fullproof}{\autoref{lma:lambda-star-bqo}}
    $S \cup L \cup \regions$ is finite, thus, by \autoref{prop:bqos}, $(\Lambda, \subseteq)$ is a \bqo.
    Again by \autoref{prop:bqos}, $(\Lambda^*, \preccurlyeq)$ is a \bqo.
    Also, again by \autoref{prop:bqos}, $(2^{\Lambda^*}, \lpowleq)$ is also a \bqo.
  \end{fullproof}
\end{lemma}

Until now, we have synthesized a \wqo for the configurations $\hset$ of a state $(s,\rho,\hset)$ of $\detdisq$, which is a Cartesian product of a symbolic trace $\symtrace$, the remaining program $\rho$ and the set of regionalized configurations $\hset$.
We use the following lemma to state that \acp{wqo} are closed under finite Cartesian products.

\begin{lemma}[\cite{kruskalWellQuasiOrderingTreeTheorem1960}]\label{lma:wqo-cartesian}
	The Cartesian product of a finite number of \acp{wqo} is a \wqo.
\end{lemma}

It remains to establish a \wqo  over symbolic traces and remaining programs to  establish an \wqo over states of $\detdisq$:

\newcommand*{\detleq}{\ensuremath{\leq_d}}
\begin{definition}
  The ordering $(\detstates, \detleq)$ between states of $\detdisq$ is defined as follows:
  $(\symtrace, \rho, \hset) \detleq (\symtrace', \rho', \hset')$ iff
  \begin{enumerate*}[label=(\arabic*)]
    \item $F(\tw(\symtrace)) = F(\tw(\symtrace'))$,
    \item $\rho = \rho'$, and
    \item $\hset \lpowleq \hset'$.
  \end{enumerate*}
\end{definition}

\begin{theorem}\label{thm:det-ordering-is-wqo}
  $(\detstates, \detleq)$ is a \wqo.
  \begin{fullproof}{\autoref{thm:det-ordering-is-wqo}}
    For $F(\symtrace) = F(\symtrace')$ as well as $\rho = \rho'$, note that the sets $\primes$ and $\sub(\delta)$ are finite.
    Thus, by \autoref{prop:bqos}, the orderings $(2^{\primes}, \subseteq)$ and $(2^{\sub(\delta)}, \subseteq)$ and thus also $(2^{\primes}, =)$ and $(2^{\sub(\delta)}, =)$ are \bqo{}s.
    Furthermore, by \autoref{lma:lambda-star-bqo}, $(2^{\Lambda^*}, \lpowleq)$ is a \bqo.
    Thus, by \autoref{lma:wqo-cartesian} and because each \bqo is also a \wqo, $\detleq$ is a \wqo.
  \end{fullproof}
\end{theorem}

With this theorem, we have defined a \wqo over states of $\detdisq$.
This allows us to traverse \detdisq and stop expanding a branch whenever we found a node $(\symtrace, \rho, \regconf)$ with an ancestor $(\symtrace', \rho', \regconf')$ such that $(\symtrace', \rho', \regconf') \detleq (\symtrace, \rho, \regconf)$.
As $(\detdisq, \detleq)$ is a \wqo, each sub-branch will only be expanded finitely many times. Thus:
\begin{theorem}\label{thm:controller-decidable}
  The \golog controller synthesis problem for \mtl constraints (as defined in \autoref{def:control-problem}) is decidable.
  \begin{fullproof}[Proof Sketch for]{\autoref{thm:controller-decidable}}
    Following the approach presented in~\cite{bouyerControllerSynthesisMTL2006}, we explore the search tree over configurations of $\detdisq$, starting from the initial configuration $c_0$ (see \autoref{def:detdiscquotient}) to determine safe strategy in \detdisq.
    Leaf nodes in the tree are labeled \emph{bad} if they contain a bad configuration and otherwise marked as being \emph{good}.
    Intermediate nodes $(\symtrace, \rho, \regconf)$ are expanded if they do not have a predecessor $(\symtrace', \rho', \regconf')$ in the tree where $(\symtrace', \rho', \regconf') \detleq (\symtrace, \rho, \regconf)$,
    in which case they are labeled as \emph{good}.
    Labels for intermediate, expanded nodes are determined based on the labels of their child nodes: if all child-nodes are labelled \emph{good}, the node is labelled as \emph{good}, otherwise it is labelled as \emph{bad}.
    A safe strategy in \detdisq exists if the root node is labelled \emph{good}.
    Each path in the resulting search tree is finite, as we would otherwise obtain an infinite anti-chain wrt $(\detdisq, \detleq)$, a contradiction to $(\detdisq, \detleq)$ being a \wqo.
    With \autoref{thm:winning-strategy}, we obtain a decidable procedure for the controller synthesis problem.
  \end{fullproof}
\end{theorem}

\section{Conclusion}\label{sec:conclusion}
High-level control of robots is challenging, as the developer needs to take care both of the high-level behavior and the low-level details of the robot platform, which often poses implicit constraints on the program.
To alleviate this issue, we proposed to make those constraints explicit as a \acf{MTL} specification.
We have presented a theoretical framework to synthesize a \golog controller that ensures that the given specification is satisfied.
Based on an extension of \golog with clocks and adapting well-known results from \acl{TA} synthesis, we have described an effective synthesis algorithm that is guaranteed to terminate.

For future work, we plan to extend our \ac{MTL} synthesis tool \tacos~\cite{hofmannTACoSToolMTL2021} to \golog programs. 
Additionally, while \ac{MTL} over infinite words is generally undecidable, it may be interesting to restrict the specification to \emph{Safety \ac{MTL}}, which is decidable on infinite words and thus may allow controller synthesis for non-terminating \golog programs.

\bibliographystyle{kr}
\bibliography{gocos}

\ifthenelse{\boolean{proofappendix}}{
\section*{Proofs}
\includecollection{proofs}
}{}

\end{document}